\documentclass{article}

\usepackage{PRIMEarxiv}

\usepackage[utf8]{inputenc} %
\usepackage[T1]{fontenc}    %
\usepackage{natbib}
\usepackage{amsmath}
\usepackage{subcaption}
\usepackage{hyperref}       %
\usepackage{url}            %
\usepackage{booktabs}       %
\usepackage{amsfonts}       %
\usepackage{nicefrac}       %
\usepackage{microtype}      %
\usepackage{lipsum}
\usepackage{fancyhdr}       %
\usepackage{graphicx}       %
\graphicspath{{media/}}     %

\title{LaMER: Language-Conditioned Multi-Expert Routing for Weakly Supervised Referring Expression Comprehension
}

\author{
  Yu Yu Chang$^{1}$, Tzu-Hsuan Weng$^{1}$, Youssef Mohamed Amin$^{2}$, Mohaned Ahmed Elsayed$^{2}$, \\
  \textbf{Hung-Hsuan Chen$^{1}$, Min-Te Sun$^{1}$, Sherine Shawky$^{2}$} \\[1ex]
  \normalfont $^{1}$National Central University, Taoyuan, Taiwan \\
  \normalfont $^{2}$Pharos University in Alexandria, Alexandria, Egypt \\[1ex]
  \normalfont\small\texttt{oscarw510gc@gmail.com, 112403057@cc.ncu.edu.tw, 202304504@pua.edu.eg, 202304056@pua.edu.eg,} \\
  \normalfont\small\texttt{hhchen1105@acm.org, msun0807@gmail.com, sherine.shawky@pua.edu.eg} \\
}

\begin{document}
\maketitle

\begin{abstract}
Referring Expression Comprehension (REC) localizes the image region
described by a natural language expression, underpinning
applications such as autonomous navigation, human-robot interaction,
and medical image analysis. Under weak supervision, REC is
challenged by the visual-linguistic semantic gap and the difficulty
of inferring region-text correspondences from coarse image-level
signals alone. Moreover, no single visual encoder captures all
complementary properties---semantic alignment, fine-grained
texture, and boundary awareness---while naively fusing multiple
pretrained encoders yields heterogeneous, hard-to-reconcile
representations. We propose LaMER, a weakly supervised REC
framework that dynamically aggregates four frozen expert encoders
through a language-conditioned Cross-Modal Routing Fusion module,
adapting expert selection to each referring expression without
instance-level bounding-box supervision. A Cell-Text Matching Head
with Negative Anchor Augmentation, trained with a multi-objective
loss combining contrastive, reconstruction, alignment, and
load-balancing terms, further stabilizes optimization. LaMER
achieves 77.67\%, 62.81\%, and 69.49\% Acc@IoU\(\geq 0.5\) on
RefCOCO, RefCOCO+, and RefCOCOg, surpassing DViN and WeakMCN by up
to 16.49 and 14.49 points, with ablations validating each component.
\end{abstract}

\section{Introduction}
\label{sec:intro}

Understanding and localizing natural language references to visual
content is a core requirement for intelligent systems operating in
the real world. Referring Expression Comprehension (REC), the task
of localizing the image region described by a natural language
expression, underpins applications such as autonomous
navigation~\citep{deruyttere2019talk2car}, human-robot
interaction~\citep{shridhar2020ingress}, and medical image
analysis~\citep{lei2023medphrase}, where systems must identify the
target object among multiple visually similar candidates.

REC remains fundamentally challenged by a visual-linguistic semantic
gap between continuous visual representations and discrete language
tokens, an absence of instance-level supervision, and the difficulty
of integrating heterogeneous visual representations. These
challenges intensify under weak supervision, where models learn
solely from image-level signals, making the quality of visual
representations the primary bottleneck: features must be both
spatially discriminative and semantically aligned with the language
query. Yet no single visual encoder simultaneously captures all
complementary visual cues---semantic alignment, fine-grained
texture, structure, and boundary awareness. Integrating multiple
specialized pretrained encoders through a Mixture-of-Experts (MoE)
paradigm~\citep{shazeer2017moe} is a promising direction, but
heterogeneous experts produce representations optimized for
different objectives and feature distributions, making naive fusion
ineffective; a learned language-conditioned router further risks
collapsing onto a small subset of experts without appropriate
load-balancing regularization.

We review existing REC methods in detail in
Section~\ref{sec:related}. In brief, fully supervised
approaches~\citep{yu2018mattnet,hu2017cmn,wang2019neighbourhood,yang2020graph,deng2021transvg,kamath2021mdetr,choi2022seqtr}
achieve strong performance but depend on costly instance-level
annotations that limit scalability~\citep{liu2019arn}, while recent
weakly supervised methods~\citep{jin2023refclip,luo2024apl}
instead pair open-vocabulary
detectors~\citep{gu2022vild,cheng2024yoloworld} with a single visual
backbone. This reliance on one encoder limits the diversity of visual characteristics captured, leaving a gap in aggregating heterogeneous visual representations under
weak supervision.

To address these limitations, we propose LaMER
(Language-Conditioned Multi-Expert Routing for Weakly Supervised
Referring Expression Comprehension), a weakly supervised REC
framework that dynamically aggregates complementary visual
representations from heterogeneous pretrained visual experts.
Instead of a single visual backbone, LaMER leverages multiple frozen
expert encoders and aggregates their outputs through a learned gating mechanism in a Cross-Modal Routing Fusion module and a
Cell-Text Matching Head with Negative Anchor Augmentation, trained
with a multi-objective optimization strategy that stabilizes expert
routing and improves grounding performance without requiring
bounding-box annotations.

Extensive experiments on RefCOCO, RefCOCO+, and RefCOCOg show that
LaMER outperforms the state-of-the-art weakly supervised methods
DViN and WeakMCN by up to 16.49 and 14.49
percentage points, with comprehensive ablations validating the
effectiveness of each proposed component. The main contributions of
this paper are summarized as follows:

\begin{itemize}
\setlength{\itemsep}{1pt}
\setlength{\parskip}{0pt}
    \item We propose LaMER, a weakly supervised referring
    expression comprehension framework that introduces a
    language-conditioned multi-expert routing mechanism for
    aggregating complementary visual representations
    from heterogeneous pretrained visual encoders.
    \item We design a \emph{Cross-Modal Routing Fusion} module and an \emph{Expert Feature Bank}, enabling learned weighted integration of heterogeneous visual representations.
    \item We introduce a \emph{Cell-Text Matching Head} with
    Negative Anchor Augmentation and optimize the framework
    using a multi-objective training strategy that stabilizes expert
    routing and improves weakly supervised grounding without
    requiring bounding-box annotations.
    \item Extensive experiments on RefCOCO,
    RefCOCO+, and RefCOCOg demonstrate
    state-of-the-art performance among weakly supervised REC methods,
    while comprehensive ablation studies verify the effectiveness of
    each proposed component.
\end{itemize}

\section{Related Work and Preliminaries}
\label{sec:related}

\subsection{Open-Vocabulary Object Detection}
Open-vocabulary object detection (OVD) extends detection beyond predefined categories to novel concepts described through natural language. Early semantic-transfer approaches were limited by weak visual-semantic alignment; large-scale vision-language pretraining (e.g., CLIP, ALIGN) improved this via contrastive learning on image-text pairs. Building on such models, ViLD~\citep{gu2022vild} distills a frozen CLIP teacher into a detector; GLIP and RegionCLIP reformulate detection as phrase grounding; YOLO-World~\citep{cheng2024yoloworld} introduces a lightweight vision-language pathway for real-time inference. Despite these advances, OVD models remain designed for category-level recognition and are insufficient for compositional expressions involving attributes, spatial relationships, and context---so directly applying OVD representations to REC remains challenging.

\subsection{Referring Expression Comprehension}
Referring Expression Comprehension (REC) aims to localize the image region described by a natural language expression, requiring models to identify the intended instance among visually similar candidates. Existing approaches divide into fully supervised and weakly supervised paradigms.

\paragraph{Fully supervised REC.}
Early REC approaches relied on modular architectures that decomposed
expressions into semantic components such as object categories,
attributes, and spatial relationships~\citep{yu2018mattnet,hu2017cmn},
while graph-based methods modeled objects as nodes and exploited
relational reasoning over scene
structures~\citep{wang2019neighbourhood,yang2020graph}. With
Transformer-based vision-language architectures, approaches such as
TransVG~\citep{deng2021transvg}, MDETR~\citep{kamath2021mdetr}, and
SeqTR~\citep{choi2022seqtr} achieved strong performance through
end-to-end cross-modal interaction and localization, but their
dependence on expensive instance-level bounding-box annotations
limits scalability to large-scale scenarios~\citep{liu2019arn}.

\paragraph{Zero-shot and weakly supervised REC.}
To reduce dependence on annotations, studies investigate weakly supervised and zero-shot REC, learning grounding from image-level supervision or transferring it from pretrained vision-language models, formulating REC as region-text matching over open-vocabulary detector proposals~\citep{gu2022vild,cheng2024yoloworld}. RefCLIP~\citep{jin2023refclip}, APL~\citep{luo2024apl}, DViN~\citep{chen2025dvin}, and WeakMCN~\citep{cheng2025weakmcn} improve weakly supervised grounding through enhanced visual-language alignment and auxiliary learning; \cite{chu2024cma} augment RefCLIP's anchor-matching backbone with cross-modality attention, while \cite{mi2024adaptivekd} distill teacher-grounder knowledge under a ground-truth-proposal regime rather than LaMER's fully automatic candidate generation. QueryMatch~\citep{chen2024querymatch} and AlignCAT~\citep{wang2025aligncat} replace anchor-level matching with query-based representations from a Mask2Former detector; \cite{he2026nrg} adopt a grounding DINO backbone with VLM-derived pseudo-labels; \cite{liu2026confidence} propose confidence-aware pseudo-label self-correction; and \cite{xu2026svcr} propose SVCR, routing multiple pretrained visual foundation models into structured relational features---the closest architectural precedent to LaMER's multi-expert routing.

However, most of these recent methods  rely on a single visual backbone,
limiting their ability to capture the diverse visual properties
complex expressions require, and lack mechanisms to dynamically
select complementary visual information per expression. LaMER
addresses this gap via a language-conditioned routing mechanism over
heterogeneous, frozen visual expert encoders, building on three
established techniques---attention, mixture-of-experts routing, and
contrastive learning---each mapped to a specific module or loss in
Section~\ref{sec:method}; a self-contained review is provided in
Appendix~\ref{app:background}.

\section{Proposed Method}
\label{sec:method}

\subsection{Overview}

LaMER couples a YOLO11x-based visual encoder~\citep{khanam2024yolov11} and a Bi-GRU language encoder with a Cross-Modal Routing Fusion module that assigns soft weights to four frozen expert encoders and dynamically mixes them with the backbone feature through a learnable residual gate, strengthening cross-modal alignment under weak supervision.

The fused representation is subsequently processed by the Cell-Text Matching Head, which performs region-text alignment through anchor-level contrastive matching. Negative Anchor Augmentation adds challenging negatives during training to improve discrimination, and LaMER is optimized with a multi-objective loss combining contrastive, reconstruction, alignment, and load-balancing objectives, enabling stable routing without instance-level bounding-box annotations.

\subsection{Research Challenges}

Beyond the visual-linguistic semantic gap and the absence of
instance-level supervision introduced in
Section~\ref{sec:intro}, realizing this design raises two further
architecture-specific challenges:

\begin{enumerate}
\setlength{\itemsep}{1pt}
\setlength{\parskip}{0pt}
    \item \textbf{Visual Expert Heterogeneity:}
    The four frozen experts---SigLIP (image-text contrastive
    learning), DINOv2 (self-supervised learning), EfficientSAM (mask
    prediction), and ConvNeXt (supervised classification)---are
    pretrained under heterogeneous objectives and exhibit
    substantially different feature distributions, making simple
    fusion strategies such as averaging or concatenation
    ineffective.

    \item \textbf{Dynamic Routing Stability:}
    Training a language-conditioned router jointly with the grounding
    objective may collapse it toward a single dominant expert;
    balancing expert utilization against the grounding objective
    requires carefully designed load-balancing losses, as overly
    strong regularization can interfere with the primary task.
\end{enumerate}

\subsection{Proposed System Architecture}

\begin{figure}[htbp]
    \centering
    \includegraphics[width=\linewidth]{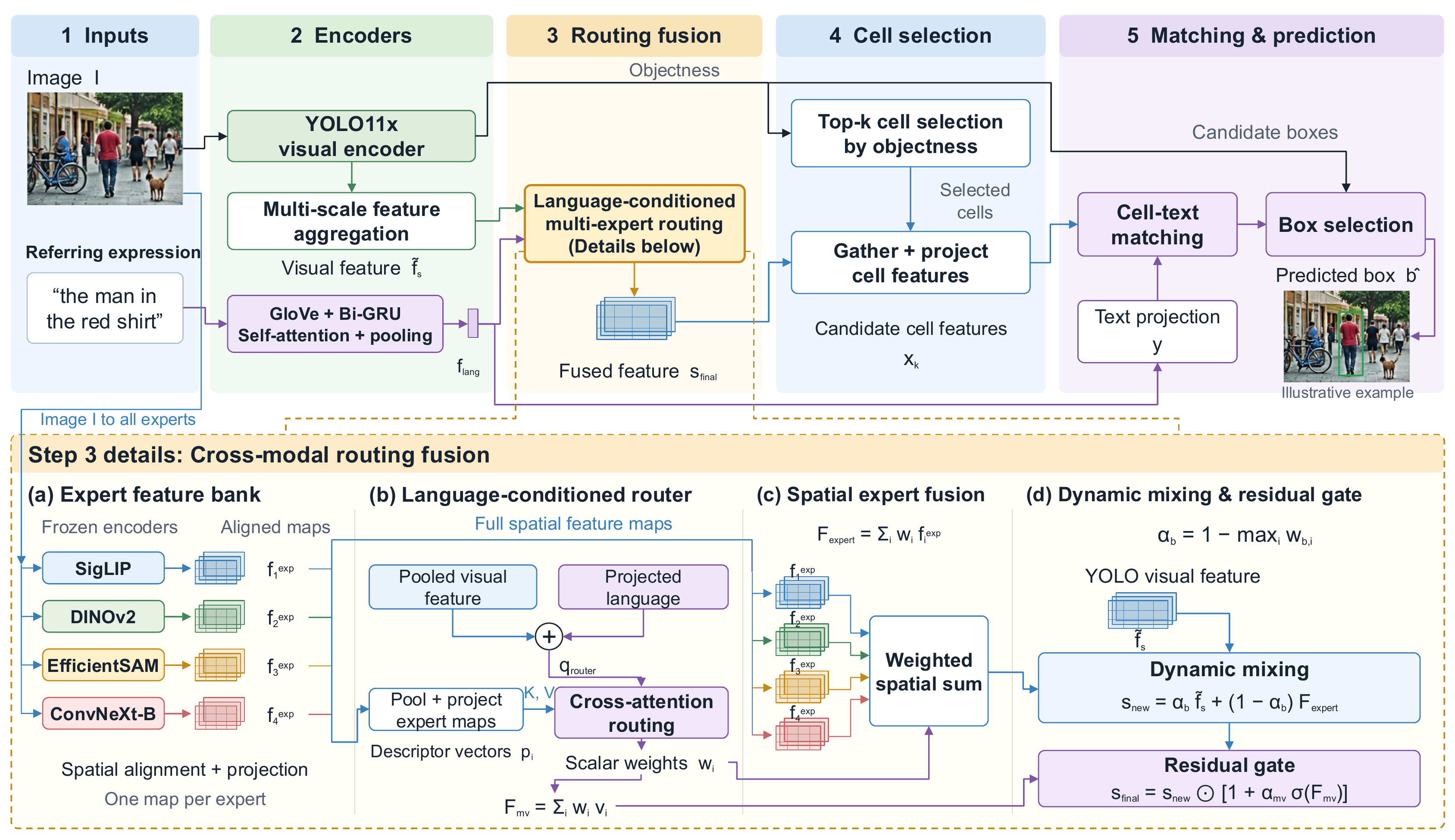}
    \caption{Overall architecture of the proposed LaMER framework.}
    \label{fig:overview}
\end{figure}

As illustrated in Figure~\ref{fig:overview}, the system consists of five components: the YOLO11x visual encoder, the Bi-GRU language encoder, the Expert Feature Bank, the Cross-Modal Routing Fusion module, and the Cell-Text Matching Head. The input image \(I \in \mathbb{R}^{H \times W \times 3}\) and referring expression \(Q = \{w_1, \ldots, w_T\}\) are fed in parallel into the visual and language encoders, extracting multi-scale visual features and a 512-dimensional context-aware language representation. These visual features are enriched by four frozen experts in the Expert Feature Bank---SigLIP SO400M~\citep{zhai2023siglip}, DINOv2 ViT-L/14~\citep{oquab2023dinov2}, EfficientSAM (ViT-T)~\citep{xiong2024efficientsam}, and ConvNeXt-B~\citep{liu2022convnext}---each projected to a common \(13
\times 13\), 1024-channel resolution. The Cross-Modal Routing Fusion module weights each expert's contribution by conditioning on a joint visual-linguistic query, producing a fused
visual feature passed to the Cell-Text Matching Head, which selects
the top-\(K\) cells by objectness, matches them against the query, and produces the final bounding box \(\hat{b} = (x, y, w,
h)\) without bounding-box supervision during training.

\subsubsection{Data Preprocessing}
\label{sec:preprocessing}

This stage converts raw image--text pairs from the RefCOCO, RefCOCO+, and RefCOCOg datasets~\citep{yu2016refcoco,mao2016refcocog,kazemzadeh2014referitgame} into a format suitable for the encoding and grounding modules; consistent with the weakly supervised setting, ground-truth boxes are withheld during training and used only for evaluation. Images are resized to \(416 \times 416\) pixels for the YOLO11x backbone, and each of the four expert encoders receives its own resolution- and normalization-matched view, since they were pretrained under different conventions; full per-branch preprocessing values are given in Appendix~\ref{app:implementation}. We tokenize each referring expression into GloVe~\citep{pennington2014glove} word-index sequences, padded or truncated to \(T_{\max} = 15\) tokens.

\subsubsection{Feature Extraction}

Three parallel branches process the preprocessed image and referring expression, producing the visual and linguistic representations that the Cross-Modal Routing Fusion module consumes.

YOLO11x~\citep{khanam2024yolov11} serves as the primary visual encoder, chosen for its ability to jointly produce region proposals and semantically rich multi-scale features within a single backbone. It builds hierarchical representations through C3k2 blocks with progressively larger receptive fields, an SPPF layer, and a C2PSA (Cross-Stage Partial with Spatial Attention) module at its final stage; the backbone is kept frozen except for C2PSA, which is selectively unfrozen so its spatial attention can adapt to the grounding task without disturbing the earlier general-purpose layers. The encoder outputs three feature maps at different spatial scales---\(f_L \in \mathbb{R}^{B \times 1024 \times 13 \times 13}\), \(f_M \in \mathbb{R}^{B \times 512 \times 26 \times 26}\), and \(f_S \in \mathbb{R}^{B \times 256 \times 52 \times 52}\)---and objectness scores from the frozen detection heads serve as the anchor candidates used in Section~\ref{sec:routing}.
The referring expression is encoded with a Bidirectional GRU (Bi-GRU) over frozen GloVe~\citep{pennington2014glove} word embeddings (300-dimension), chosen so that spatial modifiers common in referring expressions (e.g., ``on the left'') are contextualized using information from both directions rather than only after the head noun. Each direction maintains a 256-dimension hidden state; the concatenated forward/backward outputs form a 512-dimension contextual token representation, further refined by three layers of multi-head self-attention and compressed by attention-based pooling into a global language feature \(f_{\text{lang}} \in \mathbb{R}^{B \times 512}\).
Following preprocessing (Section~\ref{sec:preprocessing}), each of the four frozen experts in the Expert Feature Bank — SigLIP SO400M~\citep{zhai2023siglip}, DINOv2 ViT-L/14~\citep{oquab2023dinov2}, EfficientSAM (ViT-T)~\citep{xiong2024efficientsam}, and ConvNeXt-B~\citep{liu2022convnext} — produces a complementary representation covering cross-modal semantic alignment, fine-grained self-supervised structure, boundary-aware localization, and local texture, respectively, each projected to 1024 channels and spatially aligned to \(13 \times 13\) to match \(f_L\).

\subsubsection{Cross-Modal Routing Fusion}
\label{sec:routing}

The Cross-Modal Routing Fusion module integrates the multi-scale YOLO11x features with the four expert representations under a joint visual-linguistic query, operating in four stages: multi-modal fusion, cross-attention routing, expert fusion with dynamic mixing, and top-k cell selection.

\paragraph{Multi-Modal Fusion.} Before routing, a grounding projection adapts the YOLO11x feature maps to the task: \(f_L\) passes through a bottleneck projection (\(1024 \to 512 \to 512 \to 1024\)) to remap the C2PSA spatial attention features, while the medium- and fine-scale features pass through identity-initialized \(1 \times 1\) convolutions for stable scale-specific adjustment. The three projected maps are integrated by a custom multi-scale fusion module, \textit{MultiScaleFusion}, following a Feature Pyramid Network design~\citep{lin2017fpn}, combining an upsampling path that propagates high-level semantics to finer scales with a downsampling path that aggregates fine-grained detail into coarser scales:
\begin{equation}
\tilde{f}_s, \tilde{f}_m, \tilde{f}_l = \text{MultiScaleFusion}(f_L, f_M, f_S),
\label{eq:multiscale}
\end{equation}
where \(\tilde{f}_s \in \mathbb{R}^{B \times 1024 \times 13 \times 13}\) serves as the primary visual representation for routing. A joint visual-linguistic router query is then formed by adding the globally pooled \(\tilde{f}_s\) to the projected language feature:
\begin{equation}
q_{\text{router}} = \text{AvgPool}(\tilde{f}_s) + W_{\text{lang}} f_{\text{lang}} \in \mathbb{R}^{B \times 1024},
\label{eq:router-query}
\end{equation}
where \(W_{\text{lang}} \in \mathbb{R}^{512 \times 1024}\) projects the language feature into the routing space.

\paragraph{Cross-Attention Router.} Since a static fusion strategy cannot reflect the differing usefulness of the 4 experts for grounding, a soft routing mechanism based on scaled dot-product attention learns continuous weights to all four experts simultaneously, keeping every expert active rather than committing to a hard selection. Each expert's feature map is globally pooled and projected into a shared descriptor \(p_i \in \mathbb{R}^{B \times 1024}\), stacked into \(P \in \mathbb{R}^{B \times 4 \times 1024}\). With projections \(q = W_Q q_{\text{router}}\), \(K = W_K P\), \(V = W_V P\), the routing weights and mixed-value representation follow scaled dot-product attention:
\begin{equation}
w = \text{softmax}\!\left(\frac{qK^{\top}}{\sqrt{d}}\right) \in \mathbb{R}^{B \times 4}, \quad
F_{mv} = \sum_{i=1}^{4} w_i \cdot v_i \in \mathbb{R}^{B \times d},
\label{eq:router-attn}
\end{equation}
where \(w_i\) is the \(i\)-th expert's routing weight, and \(F_{mv}\) encodes expert selection and value blending.

\paragraph{Expert Fusion and Dynamic Mixing.} The routing weights are applied to the full spatial feature maps of the four experts to produce a weighted expert representation \(F_{\text{expert}} = \sum_{i=1}^{4} w_i \cdot f_i^{\text{exp}}\). This is blended with the YOLO11x backbone feature through a per-sample mixing coefficient derived from the peak routing weight:
\begin{equation}
\alpha_b = 1 - \max_i w_{b,i}, \qquad
s_{\text{new}} = \alpha_b \cdot \tilde{f}_s + (1 - \alpha_b) \cdot F_{\text{expert}},
\label{eq:mixing}
\end{equation}
so that a concentrated routing distribution shifts the blend toward the dominant expert, while a flatter distribution retains more of the YOLO11x backbone feature. A learnable residual gate conditioned on \(F_{mv}\) then applies a final channel-wise adjustment:
\begin{equation}
s_{\text{final}} = s_{\text{new}} \cdot \left(1 + \alpha_{mv} \cdot \sigma(F_{mv})\right),
\label{eq:residual-gate}
\end{equation}
where \(\alpha_{mv}\) is a scalar initialized to zero, so the gate begins as an identity mapping and only gradually contributes as training progresses, and \(\sigma\) is the sigmoid function.

\paragraph{Top-k Cell Selection.} After fusion, the \(13 \times 13 = 169\) grid cells are ranked by the mean objectness of their four anchor proposals from the YOLO11x detection head (\(13 \times 13 \times 4 = 676\) proposals in total), and the top \(k = 17\) cells are selected:
\begin{equation}
A_{\text{sel}} = \text{TopK}\left(\{\bar{o}_c\}_{c=1}^{169}\right),
\label{eq:topk}
\end{equation}
where \(\bar{o}_c\) is the mean objectness of the anchors in cell \(c\) and \(A_{\text{sel}}\) is the set of selected cells. The spatial features at the corresponding grid positions in \(s_{\text{final}}\) are extracted and projected to 512 dimensions, yielding \(X \in \mathbb{R}^{B \times 17 \times 512}\), while the language feature \(f_{\text{lang}}\) is projected to \(Y \in \mathbb{R}^{B \times 1 \times 512}\). The pair \((X, Y)\) is then passed to the Cell-Text Matching Head.

\subsubsection{Cell-Text Matching Head}

The Cell-Text Matching Head receives the top-\(k\) anchor feature matrix \(X \in \mathbb{R}^{B \times 17 \times 512}\) and the global language feature \(Y \in \mathbb{R}^{B \times 1 \times 512}\) from Top-k Cell Selection, and produces the final grounding prediction through contrastive cell-text matching.

A pairwise similarity score is first computed between each of the 17 anchor representations and the language query via scaled dot-product:
\begin{equation}
S = \frac{XY^{\top}}{\sqrt{d}} \in \mathbb{R}^{B \times 17},
\label{eq:matching-score}
\end{equation}
where \(d = 512\). To increase discriminative difficulty, Negative Anchor Augmentation appends the top-2 highest-scoring anchors from other samples within the same batch as hard negatives, forcing the model to distinguish the target from semantically plausible candidates drawn from entirely different images rather than trivially low-scoring distractors.

At inference, the prediction is the anchor with the highest similarity score:
\begin{equation}
\hat{k} = \arg\max_{k \in \{1, \ldots, 17\}} S_k,
\label{eq:matching-argmax}
\end{equation}
whose coordinates \((x, y, w, h)\) are decoded from the YOLO11x detection head as the final predicted bounding box \(\hat{b}\).

\subsubsection{Loss Function}

The primary weakly supervised signal is the anchor-prompt contrastive loss \(\mathcal{L}_{\text{anchor}}\), aligning the top-17 anchor visual features with the language query via the augmented InfoNCE-style~\citep{oord2018cpc} objective above.

The reconstruction loss \(\mathcal{L}_{\text{recon}}\) adds cross-modal regularization: a three-layer MLP decoder reconstructs the mean of the top-17 anchor visual features back into the language feature space:
\begin{equation}
\mathcal{L}_{\text{recon}} = \Bigl\| \frac{1}{17} \sum_{k=1}^{17} \text{Decoder}(x_k) - f_{\text{lang}} \Bigr\|_2^2,
\label{eq:recon-loss}
\end{equation}
encouraging selected anchors to retain semantic consistency with the query even after expert fusion.

To discourage expert collapse, a load-balancing loss \(\mathcal{L}_{\text{sparse}}\) penalizes deviation of the batch-averaged routing distribution from a uniform prior:
\begin{equation}
\mathcal{L}_{\text{sparse}} = \text{MSE}\!\left(\bar{w}, \tfrac{1}{4}\mathbf{1}\right), \qquad
\bar{w} = \frac{1}{B} \sum_{b=1}^{B} w_b,
\label{eq:sparse-loss}
\end{equation}
where \(w_b \in \mathbb{R}^4\) is the \(b\)-th sample's routing weight vector.

The text-visual alignment loss \(\mathcal{L}_{\text{align}}\) gives the routing module a more direct cross-modal signal, applying InfoNCE~\citep{oord2018cpc} at \(\tau = 0.5\) between projections of the fused expert features and the language feature:
\begin{equation}
\mathcal{L}_{\text{align}} = \text{InfoNCE}\left(\text{Proj}_{\text{expert}}(F_{\text{expert}}), \text{Proj}_{\text{text}}(f_{\text{lang}}); \tau = 0.5\right),
\label{eq:align-loss}
\end{equation}
where both projectors are two-layer MLPs mapping to \(\mathbb{R}^{128}\); without this loss the router would receive cross-modal gradient only indirectly through \(\mathcal{L}_{\text{anchor}}\).

Finally, the box regression loss \(\mathcal{L}_{\text{box}}\) provides optional spatial supervision: a soft-attention weighted average of the selected anchors initializes a prediction refined by a two-layer MLP residual head, supervised with Generalized IoU~\citep{rezatofighi2019giou}:
\begin{equation}
\mathcal{L}_{\text{box}} = \mathcal{L}_{\text{GIoU}}(\hat{b}_{\text{pred}}, b_{\text{gt}}).
\label{eq:box-loss}
\end{equation}

The overall training objective is a weighted sum of the five terms:
\begin{equation}
\mathcal{L}_{\text{total}} = \mathcal{L}_{\text{anchor}} + \mathcal{L}_{\text{recon}} + \lambda_{\text{sparse}} \mathcal{L}_{\text{sparse}} + \lambda_{\text{align}} \mathcal{L}_{\text{align}} + \lambda_{\text{box}} \mathcal{L}_{\text{box}}.
\label{eq:total-loss}
\end{equation}
\(\mathcal{L}_{\text{box}}\) and its refinement head support optional supervised fine-tuning, but we set \(\lambda_{\text{box}} = 0\) in all reported experiments, so ground-truth boxes never contribute gradient signal, and predictions are decoded directly from the selected anchor, consistent with the weakly supervised setting.

\section{Experiments}
\label{sec:experiments}

\subsection{Datasets}
We evaluate on three standard REC benchmarks: RefCOCO and RefCOCO+~\citep{yu2016refcoco}, both built on MS-COCO train2014 images via the interactive ReferItGame protocol~\citep{kazemzadeh2014referitgame} and split following the UNC convention (testA: person queries, testB: non-person queries), and RefCOCOg~\citep{mao2016refcocog}, collected non-interactively via Amazon Mechanical Turk~\citep{aws2005mturk} with longer, more descriptive expressions (8.4 words on average vs.\ 3.5 for RefCOCO), split following the UMD convention. RefCOCO+ additionally prohibits spatial terms (e.g., ``left''/``right''), isolating appearance-based grounding. Table~\ref{tab:dataset-stats} summarizes dataset statistics; Appendix~\ref{app:implementation} provides full per-dataset details.

\begin{table}[tb]
\centering
\caption{Statistics of the benchmark datasets.}
\label{tab:dataset-stats}
\small
\setlength{\tabcolsep}{3pt}
\begin{tabular}{lcccc}
\toprule
Dataset & Images & Objects & Expressions & Train / Val / Test \\
\midrule
RefCOCO  & 19,994 & 50,000 & 142,210 & 120.6k / 10.8k / 10.7k \\
RefCOCO+ & 19,992 & 49,856 & 141,564 & 120.1k / 10.7k / 10.6k \\
RefCOCOg & 25,799 & 49,822 & 95,010  & 80.5k / 4.8k / 9.6k \\
\bottomrule
\end{tabular}
\end{table}

\subsection{Evaluation Metrics}

Following standard practice in REC evaluation, we adopt Accuracy at IoU threshold 0.5 (Acc@IoU\(\geq 0.5\)) as the evaluation metric: a prediction is correct if the Intersection over Union between the predicted box \(\hat{b}\) and ground-truth box \(b^*\) exceeds 0.5,
\begin{equation}
\text{IoU}(\hat{b}, b^*) = \frac{\text{Area}(\hat{b} \cap b^*)}{\text{Area}(\hat{b} \cup b^*)},
\label{eq:iou}
\end{equation}
and the overall metric is the fraction of the \(N\) test samples satisfying this threshold.

\subsection{Experimental Setup}

LaMER is optimized with Adam~\citep{kingma2015adam} under a warmup-cosine schedule for 55 epochs (batch size 32, gradient accumulation over 2 steps), with the YOLO11x backbone and GloVe embeddings frozen except for the C2PSA attention block, which is selectively fine-tuned. Full hyperparameters and hardware configuration are provided in Appendix~\ref{app:implementation}.

\subsection{Experimental Results and Analysis}

\subsubsection{Comparison with State-of-the-Art Methods}

\begin{table}[tb]
\centering
\caption{Comparison with state-of-the-art methods on REC datasets.}
\label{tab:sota-comparison}
\setlength{\tabcolsep}{2pt}
\renewcommand{\arraystretch}{0.8}
\begin{tabular}{lccccccc}
\toprule
& \multicolumn{3}{c}{RefCOCO} & \multicolumn{3}{c}{RefCOCO+} & RefCOCOg \\
Method & val & testA & testB & val & testA & testB & val \\
\midrule
\multicolumn{8}{l}{\textit{GT Proposals}} \\
VC~\citep{niu2018vc} & -- & 33.3 & 30.1 & -- & 34.6 & -- & 30.3 \\
ARN~\citep{liu2019arn} & 38.1 & 36.4 & 36.5 & 34.5 & 36.4 & -- & 39.6 \\
KPRN~\citep{liu2019kprn} & 36.3 & 35.3 & 37.7 & 37.2 & 36.1 & -- & 38.4 \\
DTWREG~\citep{sun2021dtwreg} & 39.2 & 41.1 & 37.7 & 39.2 & 40.0 & -- & 43.2 \\
EARN~\citep{liu2022earn} & 38.1 & 38.3 & 38.6 & 37.5 & 37.6 & -- & 45.3 \\
Adaptive KD~\citep{mi2024adaptivekd}\(^{\dagger}\) & 39.8 & 39.9 & 40.2 & 40.1 & 40.1 & 41.1 & 48.0 \\
\midrule
\multicolumn{8}{l}{\textit{Det Proposals}} \\
VC~\citep{niu2018vc} & -- & 32.7 & 27.2 & -- & 34.7 & -- & 29.7 \\
ARN~\citep{liu2019arn} & 32.2 & 35.3 & 30.3 & 32.8 & 34.4 & -- & 33.1 \\
IGN~\citep{zhang2020ign} & 34.8 & 37.6 & 32.6 & 34.3 & 36.9 & -- & 34.9 \\
DTWREG~\citep{sun2021dtwreg} & 38.4 & 39.5 & 37.0 & 38.9 & 39.9 & -- & 42.5 \\
RefCLIP~\citep{jin2023refclip} & 60.4 & 58.6 & 57.1 & 40.4 & 40.5 & -- & 47.9 \\
QueryMatch~\citep{chen2024querymatch} & 66.0 & 66.0 & 65.5 & 44.8 & 46.7 & 41.5 & 48.5 \\
APL~\citep{luo2024apl} & 64.5 & 61.9 & 63.6 & 42.7 & 42.8 & -- & 50.2 \\
CMA~\citep{chu2024cma} & 63.9 & 61.4 & 63.8 & 41.8 & 42.1 & 40.4 & 48.4 \\
DViN~\citep{chen2025dvin} & 67.7 & 70.9 & 59.4 & 52.5 & 57.5 & -- & 55.0 \\
WeakMCN~\citep{cheng2025weakmcn} & 69.2 & 70.8 & 63.4 & 51.9 & 57.4 & 43.3 & 55.0 \\
AlignCAT~\citep{wang2025aligncat} & 69.0 & 70.3 & 66.6 & 47.2 & 52.2 & 41.9 & 54.7 \\
NRG~\citep{he2026nrg} & 69.6 & 73.2 & 67.7 & 56.0 & 64.5 & 46.2 & 58.3 \\
SVCR~\citep{xu2026svcr} & 71.1 & 72.6 & 65.6 & 54.2 & 60.2 & 44.6 & 57.0 \\
SVCR~\citep{xu2026svcr} (YOLOv5) & 74.5 & 76.9 & 70.5 & 56.4 & 64.2 & 48.7 & 60.4 \\
\midrule
\textbf{LaMER (Ours)} & \textbf{77.7} & \textbf{79.1} & \textbf{75.9} & \textbf{62.8} & \textbf{67.9} & \textbf{57.0} & \textbf{69.5} \\
\bottomrule
\end{tabular}

\vspace{0pt}
{\scriptsize \(^{\dagger}\)Adaptive KD trains on ground-truth-derived triads, an easier setting than the Det Proposals methods below; listed under GT Proposals for like-for-like grouping, not as a weaker model. LaMER's RefCOCOg result is on the UMD validation split; baseline RefCOCOg results are quoted from the original papers, and WeakMCN results from its official repository.}
\end{table}

Table~\ref{tab:sota-comparison} reports Acc@IoU\(\geq 0.5\) for LaMER against representative weakly supervised REC methods, grouped by proposal source: \textit{GT Proposals} methods rely on ground-truth region candidates at inference, while \textit{Det Proposals} methods, including LaMER, generate candidates automatically via a pretrained detector without any annotation---a harder setting, since the candidate set is not guaranteed to contain the correct region. Despite this handicap, LaMER leads all Det Proposals baselines and surpasses every GT Proposals method across all three benchmarks.

Against the strongest prior Det Proposals method, WeakMCN, LaMER's gains widen as the task grows harder: +8.45/+8.33/+12.45 on RefCOCO val/testA/testB; +10.88/+10.47/+13.73 on RefCOCO+, which prohibits spatial terms and forces appearance-based reasoning, indicating the Expert Feature Bank's texture- and structure-sensitive experts supply discriminative information single-encoder methods lack; and the largest margin overall, +14.49 over WeakMCN and +19.27 over APL, on RefCOCOg, whose longer expressions combine appearance and relational cues---suggesting the Cross-Attention Router adapts its expert weighting more effectively to multi-clause queries.

Query-based matching (QueryMatch~\citep{chen2024querymatch}, AlignCAT~\citep{wang2025aligncat}) and enhanced anchor-matching (CMA~\citep{chu2024cma}) all trail LaMER by 8 to 26 points, indicating that stronger cross-modal alignment alone does not close the gap left by a single visual backbone. NRG~\citep{he2026nrg}, fine-tuning Grounding DINO via LoRA with VLM-derived pseudo-labels, trails by 3.4 to 11.2 points. The closest competitor is SVCR~\citep{xu2026svcr}, routing multiple foundation models into relational representations: under its default YOLOv3 backbone, SVCR trails by 6.5 to 12.5 points, narrowing to +2.16 on RefCOCO testA with a stronger YOLOv5 backbone. Even so, LaMER outperforms both SVCR and NRG on every split under every backbone setting.

\subsection{Ablation Studies}

We ablate each component on RefCOCO and RefCOCO+ under Acc@IoU\(\geq 0.5\). The expert-combination and encoder-count studies are reported in Appendix~\ref{app:extra-ablations}; here we focus on the cumulative component ablation.

\subsubsection{Cumulative Component Ablation}

\begin{table}[tb]
\centering
\caption{Cumulative ablation study of LaMER.}
\label{tab:cumulative-ablation}
\small
\setlength{\tabcolsep}{3pt}
\renewcommand{\arraystretch}{0.82}
\begin{tabular}{lcccccc}
\toprule
& \multicolumn{3}{c}{RefCOCO} & \multicolumn{3}{c}{RefCOCO+} \\
Method & val & testA & testB & val & testA & testB \\
\midrule
YOLO11x + 4 Experts & 74.1 & 74.2 & 72.9 & 56.2 & 61.3 & 50.9 \\
+ Cross-Attn Router & 76.6 & 77.9 & 73.6 & 60.8 & 65.9 & 55.4 \\
+ MultiScale Fusion & 77.5 & 78.3 & 74.2 & 60.2 & 65.5 & 53.8 \\
+ Bottleneck Proj. & 77.6 & 78.7 & 75.4 & 62.3 & 67.6 & 56.2 \\
+ Unfreeze C2PSA & \textbf{77.7} & \textbf{79.1} & \textbf{75.9} & \textbf{62.8} & \textbf{67.9} & \textbf{57.0} \\
\bottomrule
\end{tabular}
\end{table}

Table~\ref{tab:cumulative-ablation} adds each architectural component incrementally. The baseline of YOLO11x with four statically fused experts reaches 74.11\% and 56.22\% on the RefCOCO and RefCOCO+ val splits. The Cross-Attention Router yields the largest single gain (+2.51 and +4.53 points), confirming that learned expert weighting is central to exploiting complementary experts. Multi-Scale Fusion further improves RefCOCO val by +0.83, though RefCOCO+ val regresses slightly (\(-0.57\)) at this stage; the Bottleneck grounding projection recovers this, adding +2.09 on RefCOCO+ val, so the two components are complementary. Selectively unfreezing C2PSA provides a final, smaller gain (+0.40/+0.30 testA), consistent with its role of limited task-specific adaptation.

\begin{figure}[tb]
    \centering
    \begin{subfigure}[t]{0.235\linewidth}
        \includegraphics[width=\linewidth]{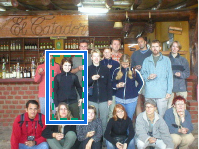}
        \caption{woman in black sweatshirt standing with brown hair}
    \end{subfigure}\hfill
    \begin{subfigure}[t]{0.235\linewidth}
        \includegraphics[width=\linewidth]{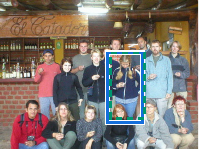}
        \caption{woman in dark blue jacket with long blond hair}
    \end{subfigure}\hfill
    \begin{subfigure}[t]{0.235\linewidth}
        \includegraphics[width=\linewidth]{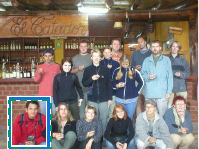}
        \caption{far left}
    \end{subfigure}\hfill
    \begin{subfigure}[t]{0.235\linewidth}
        \includegraphics[width=\linewidth]{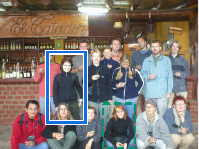}
        \caption{girl glasses black shirt front}
    \end{subfigure}

    \vspace{4pt}

    \begin{subfigure}[t]{0.235\linewidth}
        \includegraphics[width=\linewidth]{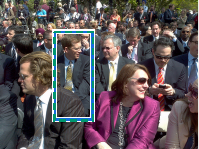}
        \caption{second row man yellow tie on the left}
    \end{subfigure}\hfill
    \begin{subfigure}[t]{0.235\linewidth}
        \includegraphics[width=\linewidth]{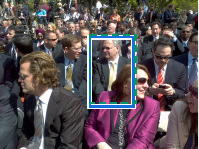}
        \caption{center man white hair yellow tie}
    \end{subfigure}\hfill
    \begin{subfigure}[t]{0.235\linewidth}
        \includegraphics[width=\linewidth]{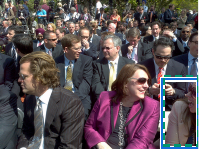}
        \caption{bottom right girl half out}
    \end{subfigure}\hfill
    \begin{subfigure}[t]{0.235\linewidth}
        \includegraphics[width=\linewidth]{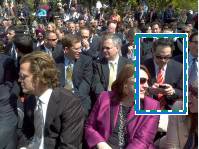}
        \caption{man right striped tie}
    \end{subfigure}

    \caption{Same image, different queries. Blue: LaMER prediction; green dashed: annotated target. Panels (a)--(d) are from a ReferItGame test image and (e)--(h) from a RefCOCO testA image. The queries in (a)--(c) and (e)--(h) lead LaMER to seven different people; in (d) the target is among the candidate boxes, but LaMER returns the box of (a).}
    \label{fig:same-image}
\end{figure}

\subsection{Qualitative Analysis}
\label{sec:qualitative-analysis}

Figure~\ref{fig:same-image} pairs two images with eight referring expressions. The predictions follow the queries: across the two scenes, the model resolves seven different people, including distinctions that rest on clothing color, hair color, relative position, and partial visibility. Panel (d) is the one failure, where the annotated target is among the candidate boxes but the model returns the box of (a).

\section{Conclusion and Future Work}
\label{sec:conclusion}
This work presents LaMER, a weakly supervised REC framework that aggregates complementary visual representations in a query-adaptive manner without instance-level supervision. It couples a YOLO11x visual encoder and a Bi-GRU language encoder with an Expert Feature Bank of four frozen encoders via a Cross-Modal Routing Fusion module and a Cell-Text Matching Head with Negative Anchor Augmentation. Trained end-to-end with a multi-objective loss, LaMER achieves 77.67\%, 62.81\%, and 69.49\% Acc@IoU\(\geq 0.5\) on RefCOCO, RefCOCO+, and RefCOCOg, surpassing DViN and WeakMCN by up to +16.49\% and +14.49\%, with ablations confirming each component's contribution. Future work includes extending LaMER to weakly supervised Referring Expression Segmentation and exploring adaptive sparse expert routing to reduce inference overhead.

\bibliographystyle{unsrt}  
\bibliography{references} 

\appendix

\section{Background}
\label{app:background}

This appendix reviews the three established techniques LaMER builds
on -- attention mechanisms, mixture-of-experts routing, and
contrastive learning -- and how each maps onto a specific module or
loss described in Section~\ref{sec:method}.

\paragraph{Attention Mechanisms.}
\label{sec:attention}
Attention mechanisms model dependencies within or between sequences
by computing relevance scores between queries and key-value
pairs, dispensing with recurrence to enable parallel
computation~\citep{vaswani2017attention}. Scaled Dot-Product Attention
computes a weighted sum of values based on query-key similarity,
while Multi-Head Attention runs several such operations in parallel
across distinct learned subspaces, allowing the model to jointly
attend to different types of relationships within the input. LaMER
uses both mechanisms directly: scaled dot-product attention forms the
basis of the Cross-Attention Router (Section~\ref{sec:routing}), and
multi-head self-attention refines the language representation
(Section~\ref{sec:preprocessing}). Several of LaMER's frozen expert
encoders also build on the closely related Vision Transformer
architecture~\citep{dosovitskiy2021image}: the Segment Anything
Model~\citep{kirillov2023sam} uses a ViT image encoder pretrained via
Masked Autoencoders~\citep{he2022mae} to produce spatially rich
embeddings, while DINO-ViT~\citep{caron2021dino} uses self-supervised
distillation to learn structural, label-free visual representations.
LaMER leverages both properties through its EfficientSAM and DINOv2
experts, respectively.

\paragraph{Mixture of Experts.}
\label{sec:moe}
To enhance model capacity and adaptability without a proportional increase in
computational cost, the Mixture of Experts (MoE) paradigm~\citep{shazeer2017moe}
is widely utilized. An MoE layer consists of a set of \(N\) expert networks
\(\{E_i\}_{i=1}^N\) and a gating network \(G\), which dynamically routes inputs
to different experts for conditional computation. For an input \(x\), the
output is a weighted sum of the experts' outputs, typically using a
Top-\(K\) routing strategy where only the \(k\) experts with the
highest gating scores are activated:

\begin{equation}
y = \sum_{i=1}^{N} G(x)_i E_i(x)
\label{eq:moe-output}
\end{equation}

\begin{equation}
G(x) = \text{Softmax}(\text{TopK}(H(x), k))
\label{eq:moe-gating}
\end{equation}

where \(G(x)_i\) indicates the gating weight for the \(i\)-th expert, \(E_i(x)\) is
the output of the \(i\)-th expert, and \(H(x)\) represents the raw logits produced
by the gating network.

\paragraph{Load Balancing.}
\label{sec:load-balancing}
During MoE training, a frequently encountered issue is expert collapse, where the gating
mechanism disproportionately assigns most inputs to a small subset of experts, resulting
in the remaining experts being rarely activated. To alleviate this imbalance and promote
more even expert participation, an auxiliary load balancing loss is introduced and defined
as follows:

\begin{equation}
L_{aux} = N \sum_{i=1}^{N} f_i \cdot P_i
\label{eq:moe-loadbalance}
\end{equation}

where \(N\) represents the total number of experts. In this formulation, \(f_i\) indicates the
empirical fraction of samples routed to expert \(i\), while \(P_i\) denotes the average routing
probability assigned to expert \(i\) by the gating network:

\begin{equation}
f_i = \frac{1}{B} \sum_{j=1}^{B} \mathbb{1}[i \in T_j]
\label{eq:moe-fi}
\end{equation}

\begin{equation}
P_i = \frac{1}{B} \sum_{j=1}^{B} G(x_j)_i
\label{eq:moe-pi}
\end{equation}

where \(T_j\) refers to the set of top-\(k\) experts chosen for input \(x_j\), \(G(x_j)_i\) is the gating
network's predicted probability of selecting expert \(i\) given \(x_j\), and \(B\) is the batch size.
By minimizing \(L_{aux}\), the model is encouraged to distribute inputs more evenly across experts,
thereby improving overall expert utilization and reducing the risk of collapse~\citep{omi2025loadbalance}.
While Eq.~\eqref{eq:moe-loadbalance} presents the canonical
load-balancing formulation, Section~\ref{sec:routing} introduces an
MSE-based variant (Eq.~\eqref{eq:sparse-loss}) tailored to LaMER's
four-expert, per-sample routing setting.

\paragraph{Contrastive Learning.}
\label{sec:contrastive}
Contrastive learning is a representation learning paradigm that structures an embedding
space by pulling semantically related sample pairs closer together while pushing apart
unrelated pairs. In multimodal learning settings, this approach is commonly used to align
latent representations across different modalities. By constructing paired samples from
different modalities or views, the model learns to map heterogeneous inputs into a shared
embedding space, enabling semantic correspondence between representations.

A widely adopted objective function for contrastive learning is the InfoNCE loss. Given
a batch containing \(N\) paired samples, the loss for the \(i\)-th pair is defined as:

\begin{equation}
L_{\text{InfoNCE}} = - \log \frac{\exp(\text{sim}(z_i^{(1)}, z_i^{(2)}) / \tau)}
{\sum_{j=1}^{N} \exp(\text{sim}(z_i^{(1)}, z_j^{(2)}) / \tau)}
\label{eq:info-nce}
\end{equation}

where \(z_i^{(1)}\) and \(z_i^{(2)}\) denote the embeddings of the two corresponding views or modalities in
the \(i\)-th paired sample, \(\text{sim}(\cdot)\) typically represents cosine similarity, and \(\tau\) is a temperature
hyperparameter that controls the smoothness of the similarity distribution. This objective
increases the similarity between matched pairs while relatively decreasing the similarity
to mismatched pairs within the batch, thereby encouraging the learning of discriminative
and well-aligned representations.

\section{Implementation Details}
\label{app:implementation}

\subsection{Dataset Details}

\textbf{RefCOCO}~\citep{yu2016refcoco} is built on the MS-COCO train2014 image collection, collected through the ReferItGame~\citep{kazemzadeh2014referitgame} platform, an interactive two-player game in which one annotator describes a target object while the other identifies it. The dataset contains 142,210 referring expressions over 50,000 annotated objects across 19,994 images, with an average expression length of 3.5 words. Following the standard UNC split, the data is divided into 120,624 training, 10,834 validation, 5,657 testA, and 5,095 testB expressions. TestA contains expressions referring exclusively to people, while testB covers all other object categories.

\textbf{RefCOCO+}~\citep{yu2016refcoco} shares the same images and collection protocol as RefCOCO but explicitly prohibits spatial location terms such as ``left'' or ``right,'' requiring annotators to rely purely on appearance-based cues such as color, texture, and category. The dataset contains 141,564 expressions over 49,856 objects across 19,992 images, with an average query length of 3.53 words, and follows the same UNC split structure as RefCOCO: 120,191 training, 10,758 validation, 5,726 testA, and 4,889 testB expressions.

\textbf{RefCOCOg}~\citep{mao2016refcocog} is also built on MS-COCO train2014 images, but is collected non-interactively through Amazon Mechanical Turk~\citep{aws2005mturk} rather than a two-player game, resulting in longer and more descriptive expressions. The dataset contains 95,010 expressions over 49,822 objects across 25,799 images , with an average expression length of 8.4 words. Unlike the other two datasets, RefCOCOg does not use a testA/testB split; following the UMD partition, it uses a single, undivided test set that ensures no image appears in more than one subset.

\subsection{Experimental Environment}

All experiments are conducted on a server equipped with two NVIDIA GeForce RTX 5090 GPUs (32 GB GDDR7 each), an AMD EPYC 9654 96-core CPU, and 256 GB of system RAM, running Ubuntu 22.04 LTS with CUDA 13.0 and PyTorch 2.11.0.

\begin{table}[htbp]
\centering
\caption{Experimental environment configuration.}
\label{tab:env-config}
\small
\setlength{\tabcolsep}{5pt}
\begin{tabular}{ll}
\toprule
Component & Specification \\
\midrule
Operating System & Ubuntu 22.04 LTS \\
CPU & AMD EPYC 9654 (96 cores) \\
System Memory & 256 GB \\
GPU & 2\(\times\) RTX 5090 (32 GB GDDR7) \\
CUDA Version & 13.0 \\
DL Framework & PyTorch 2.11.0 \\
\bottomrule
\end{tabular}
\end{table}

\subsection{Training Hyperparameters}

The model is optimized using Adam~\citep{kingma2015adam} (\(\beta_1 = 0.9\), \(\beta_2 = 0.98\), \(\epsilon = 10^{-9}\)) with a warmup cosine annealing learning rate schedule: a 5-epoch warmup from \(10^{-7}\), a peak learning rate of \(2 \times 10^{-4}\), and a minimum of \(10^{-6}\). Training runs for 55 epochs, determined empirically via validation performance on RefCOCO, with a total batch size of 32 (16 per GPU) and  gradient accumulation over 2 steps (effective batch size 64). The maximum query length is 15 tokens for RefCOCO and RefCOCO+, and 20 for RefCOCOg. Gradient clipping (max norm 0.15) and an Exponential Moving Average (EMA) of model weights are applied to stabilize and improve generalization, and automatic mixed precision (AMP) is used throughout. The YOLO11x backbone and GloVe embeddings are initialized from pretrained weights and kept frozen, with the exception of the C2PSA attention block at layer 10 of the YOLO11x backbone, which is partially fine-tuned to adapt to the grounding task.

\subsection{Per-Branch Preprocessing }

Input images are resized to \(416 \times 416\) pixels and normalized with ImageNet statistics (mean \(= [0.485, 0.456, 0.406]\), std \(= [0.229, 0.224, 0.225]\)). Because the four expert encoders in the Expert Feature Bank were pretrained under different resolutions and normalization conventions, each requires its own preprocessing branch: SigLIP SO400M receives \(384 \times 384\) input, re-normalized to its own statistics (mean \(=\) std \(= [0.5, 0.5, 0.5]\)); with a \(14 \times 14\) patch size, this yields \(27 \times 27 = 729\) patch tokens, later pooled to the target \(13 \times 13\) grid. DINOv2 ViT-L/14 receives \(364 \times 364\) input under standard ImageNet normalization. EfficientSAM (ViT-T) receives \(1024 \times 1024\) input, also under ImageNet normalization. ConvNeXt-B operates directly on the same \(416 \times 416\) ImageNet-normalized image as YOLO, requiring no additional resizing, since its cumulative \(32\times\) downsampling stride natively produces \(13 \times 13\) feature maps. Each referring expression is tokenized and mapped to GloVe~\citep{pennington2014glove} word-index sequences, padded or truncated to a maximum length of \(T_{\max} = 15\) tokens (20 for RefCOCOg).

\section{Additional Ablation Studies}
\label{app:extra-ablations}

This appendix reports two additional ablation studies referenced in
Section~\ref{sec:experiments}, both conducted on RefCOCO and
RefCOCO+ under Acc@IoU\(\geq 0.5\).

\subsection{Effect of Expert Feature Combinations}

\begin{table}[htbp]
\centering
\caption{Comparison of different mixtures of visual features on RefCOCO and RefCOCO+.}
\label{tab:expert-combinations}
\small
\setlength{\tabcolsep}{3pt}
\begin{tabular}{lcccccc}
\toprule
& \multicolumn{3}{c}{RefCOCO} & \multicolumn{3}{c}{RefCOCO+} \\
Expert Encoder & val & testA & testB & val & testA & testB \\
\midrule
YOLO11x (Base) & 67.2 & 65.4 & 70.9 & 38.1 & 37.7 & 39.4 \\
CLIP & 37.5 & 44.1 & 30.2 & 36.6 & 41.8 & 27.0 \\
SigLIP-ViT & 49.3 & 55.7 & 46.2 & 39.7 & 46.6 & 31.4 \\
DINOv2 ViT-B & 59.1 & 59.5 & 56.3 & 40.5 & 48.0 & 34.1 \\
DINOv2 ViT-L & 60.3 & 61.6 & 59.6 & 42.5 & 49.0 & 35.8 \\
ConvNeXt-B & 29.4 & 34.7 & 24.0 & 29.2 & 34.2 & 24.0 \\
EffSAM-ViT & 45.9 & 48.6 & 45.4 & 31.5 & 34.7 & 29.5 \\
\midrule
+ DINOv2 & 74.2 & 75.8 & 75.0 & 56.5 & 59.1 & 52.0 \\
+ DINO + Sig & 76.4 & 77.3 & 75.6 & 58.4 & 62.9 & 53.3 \\
+ D + Sig + SAM & 76.9 & 78.4 & 75.8 & 59.5 & 62.8 & 54.1 \\
\midrule
\textbf{LaMER (Ours)} & \textbf{77.7} & \textbf{79.1} & \textbf{75.9} & \textbf{62.8} & \textbf{67.9} & \textbf{57.0} \\
\bottomrule
\end{tabular}
\end{table}

Table~\ref{tab:expert-combinations} evaluates individual and combined expert configurations. Used in isolation, every expert encoder underperforms the YOLO11x baseline on RefCOCO (67.21\% val), with DINOv2 ViT-L achieving the strongest single-encoder result at 60.30\%; CLIP alone yields only 37.51\%, indicating that a single cross-modal encoder is insufficient for the fine-grained localization required under weak supervision. Progressively combining experts with YOLO11x yields consistent, substantial gains: adding DINOv2 ViT-L alone raises RefCOCO val from 67.21\% to 74.23\%, and subsequent additions of SigLIP and EfficientSAM further improve performance to 76.37\% and 76.87\%. Including ConvNeXt-B as the fourth expert reaches the full LaMER configuration at 77.67\% on RefCOCO val and 62.81\% on RefCOCO+ val, indicating that each expert contributes incremental, complementary information rather than redundant signal.

\subsection{Ablation of Visual Encoder Count}

\begin{table}[htbp]
\centering
\caption{Ablation of the number of visual experts.}
\label{tab:encoder-count}
\small
\setlength{\tabcolsep}{3pt}
\begin{tabular}{lcccccc}
\toprule
& \multicolumn{3}{c}{RefCOCO} & \multicolumn{3}{c}{RefCOCO+} \\
Top-\(N\) & val & testA & testB & val & testA & testB \\
\midrule
Top-1 & 38.1 & 38.6 & 38.3 & 29.2 & 29.2 & 26.6 \\
Top-2 & 75.5 & 76.4 & 74.0 & 59.3 & 64.9 & 53.2 \\
Top-3 & 77.1 & 78.8 & 75.5 & 61.0 & 67.5 & 55.1 \\
Top-4 & \textbf{77.7} & \textbf{79.1} & \textbf{75.9} & \textbf{62.8} & \textbf{67.9} & \textbf{57.0} \\
\bottomrule
\end{tabular}
\end{table}

Table~\ref{tab:encoder-count} varies the number of experts retained by the router at inference. Restricting the model to a single expert collapses performance to 38.06\% and 29.15\% on RefCOCO and RefCOCO+ val, far below any multi-expert configuration, confirming that cross-expert complementarity, rather than any individual encoder, drives the routing mechanism's effectiveness. Expanding to Top-2 recovers most of the gap (75.48\% and 59.27\%), and performance continues to improve monotonically through Top-3 and Top-4, with larger gains on RefCOCO+ than RefCOCO from Top-2 onward. This monotonic trend indicates that soft attention-based routing continues to extract complementary information as more experts are made available, without the diminishing returns typically associated with hard, single-expert selection.

\subsection{Routing Weight Analysis}
\label{app:routing}

This appendix reports the routing weights behind the qualitative results of Section~\ref{sec:qualitative-analysis} and measures how much they contribute to the predictions.

\begin{table}[htbp]
\centering
\caption{IoU with the annotated target and router weights over the four experts, for panels (a)--(h) of Figure~\ref{fig:same-image}.}
\label{tab:same-image}
\small
\begin{tabular}{@{}l*{8}{c}@{}}
\toprule
 & (a) & (b) & (c) & (d) & (e) & (f) & (g) & (h) \\
\midrule
IoU & 0.72 & 0.82 & 0.90 & 0.00 & 0.97 & 0.87 & 0.83 & 0.90 \\
\midrule
SigLIP & 0.29 & 0.31 & 0.27 & 0.30 & 0.28 & 0.26 & 0.27 & 0.26 \\
DINOv2 & 0.17 & 0.10 & 0.29 & 0.15 & 0.18 & 0.14 & 0.25 & 0.14 \\
EfficientSAM & 0.49 & 0.53 & 0.31 & 0.50 & 0.50 & 0.58 & 0.44 & 0.57 \\
ConvNeXt & 0.05 & 0.06 & 0.12 & 0.05 & 0.04 & 0.02 & 0.04 & 0.03 \\
\bottomrule
\end{tabular}

\end{table}

Table~\ref{tab:same-image} gives the weights for the eight panels of Figure~\ref{fig:same-image}. They stay within narrow bands---SigLIP $[0.26, 0.31]$, DINOv2 $[0.10, 0.29]$, EfficientSAM $[0.31, 0.58]$, ConvNeXt $[0.02, 0.12]$---averaging $[0.28, 0.18, 0.49, 0.05]$, and EfficientSAM is the highest-weighted expert in all eight. Two quantities inherit this stability: the mixing coefficient $\alpha_b = 1 - \max_i w_i$ of Eq.~\eqref{eq:mixing} stays in $[0.42, 0.69]$, and the mean utilization is far from the uniform prior of Eq.~\eqref{eq:sparse-loss}, so $\mathcal{L}_{\text{sparse}}$ moderates the imbalance without removing it, while still keeping every expert active.

\end{document}